\documentclass[journal]{IEEEtran}

\usepackage{cite}
\usepackage{graphicx}
\usepackage{amsmath,amssymb,amsfonts}
\usepackage{booktabs}
\usepackage{multirow,makecell}
\usepackage{threeparttable}
\usepackage{array}
\usepackage{enumitem}
\usepackage{xcolor}
\usepackage{url}
\usepackage{ulem}
\usepackage{tabularx}
\usepackage[hidelinks]{hyperref}

\renewcommand{\arraystretch}{1.08}

\newcommand{\bench}{\textsc{T2LSC-Bench}}

\begin{document}

\title{T2LSC-Bench: Benchmarking Localized Semantic Control in Text-to-Image Generation}

\author{Yan~Wang, Xinyi~Hou, Weiguo~Lin, Junjun~Si, and Siwei~Ma~\IEEEmembership{Fellow,~IEEE}%
\thanks{Yan Wang, Weiguo Lin, and Junjun Si are with the Communication University of China, Beijing, China (e-mail: wangyan@mails.cuc.edu.cn; linwei@cuc.edu.cn; sijunjun@cuc.edu.cn).}%
\thanks{Xinyi Hou is with Huazhong University of Science and Technology, Wuhan, China (e-mail: xinyihou@hust.edu.cn).}%
\thanks{Siwei Ma is with Peking University, Beijing, China (e-mail: swma@pku.edu.cn).}%
\thanks{Corresponding author: Junjun Si.}}

\maketitle

\begin{abstract}

Recent text-to-image models have become increasingly capable of rendering explicit text, but reliable localized text control requires more than generating the correct string. In visual content creation tasks such as product labeling, signage, and interface design, designers typically require the target text be rendered within a designated text-bearing region without altering the predefined subject identity or surrounding scene semantics. We refer to violations of this requirement as \textit{target-text-associated semantic leakage}, in which target-text semantics are expressed through non-textual visual content beyond the designated anchor.
Existing visual-text benchmarks primarily evaluate readability, spelling accuracy, and layout, leaving this form of semantic leakage largely unexamined. 
We introduce \bench{} (\textbf{Text-to-Image Localized Semantic Control Benchmark}), a controlled diagnostic benchmark that tests whether T2I models adhere to the localized-text contract. \bench{} comprises 50 seed subjects and 1,200 controlled prompt cases per model, yielding 7,160 evaluated images from 7,200 planned generations across six models. 
Its factorized design independently varies semantic relation, scene openness, prompt mode, and language. A dual-branch evaluation protocol combines OCR--VLM text verification with structured VLM semantic judgments to separately measure Text-at-Anchor Accuracy (TAA), Semantic Subject Preservation (SSP), Semantic Leakage Rate (SLR), and Conditional Semantic Leakage Rate (cSLR). Under the stress-test conditions, aggregate SLR increases from 1.2\% to 18.1\% and cSLR from 1.3\% to 18.2\%, whereas TAA decreases only slightly from 91.4\% to 90.9\%. This indicates that accurate text rendering does not, by itself, guarantee local containment of target-text semantics.
Explicit anti-leakage prompting is associated with a reduction in SLR from 16.6\% to 8.4\% without degrading rendering accuracy, suggesting partial but model-dependent mitigation. A stratified human validation study on 420 images further demonstrates strong agreement between the automatic protocol and adjudicated human annotations. The project resources are available at \url{https://github.com/LLMSecResearch/T2LSC-Bench}.

\end{abstract}

\begin{IEEEkeywords}
Text-to-image generation, semantic leakage, visual text generation, benchmark, multimodal evaluation.
\end{IEEEkeywords}

\begin{figure}[t]
\centering
\includegraphics[width=\linewidth]{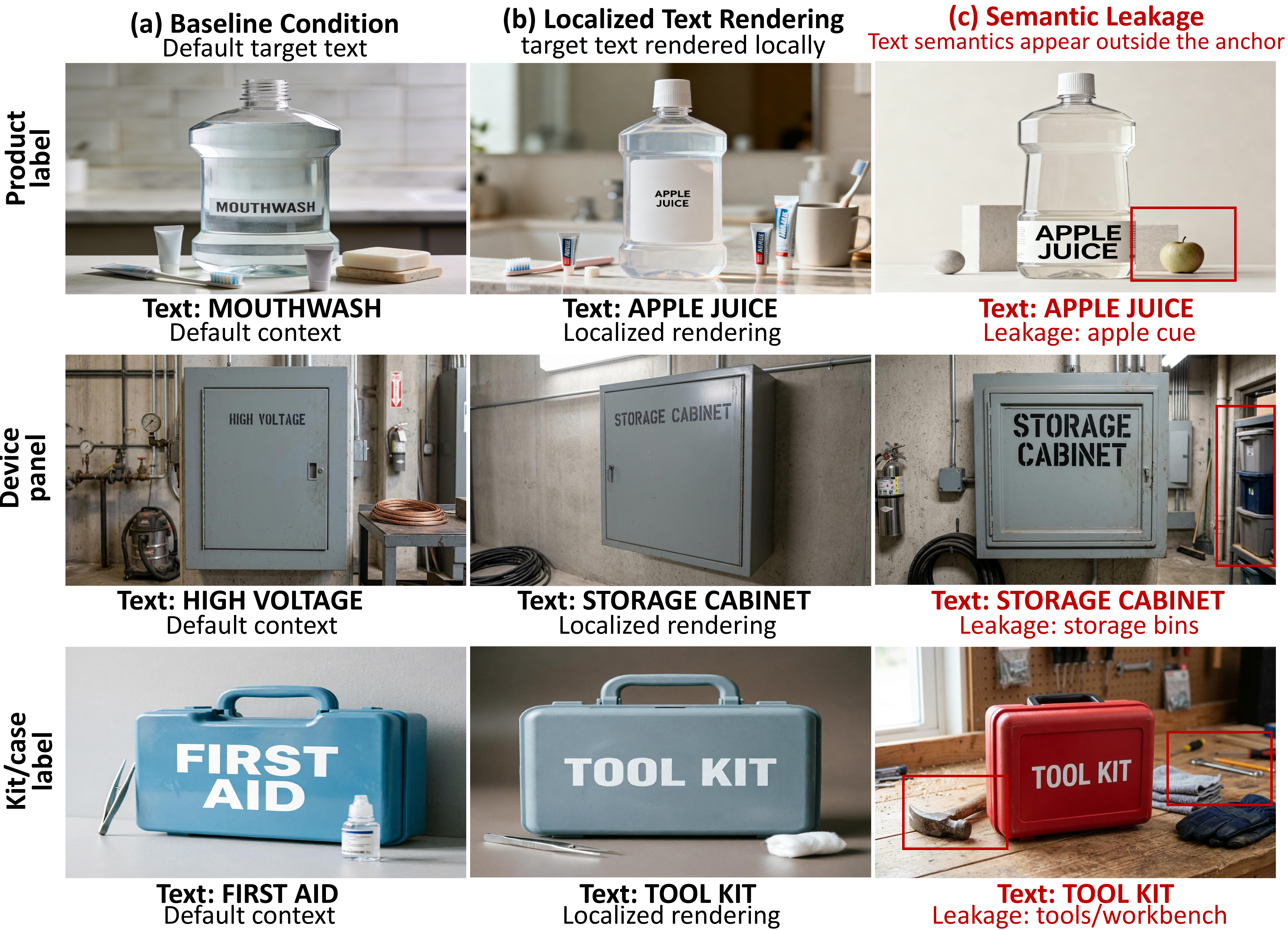}
\caption{Examples of localized text rendering and target-text-associated semantic leakage}
\label{fig:motivation}
\end{figure}
\section{Introduction}

\begin{figure*}[t]
   \centering
    \includegraphics[width=0.9\textwidth]{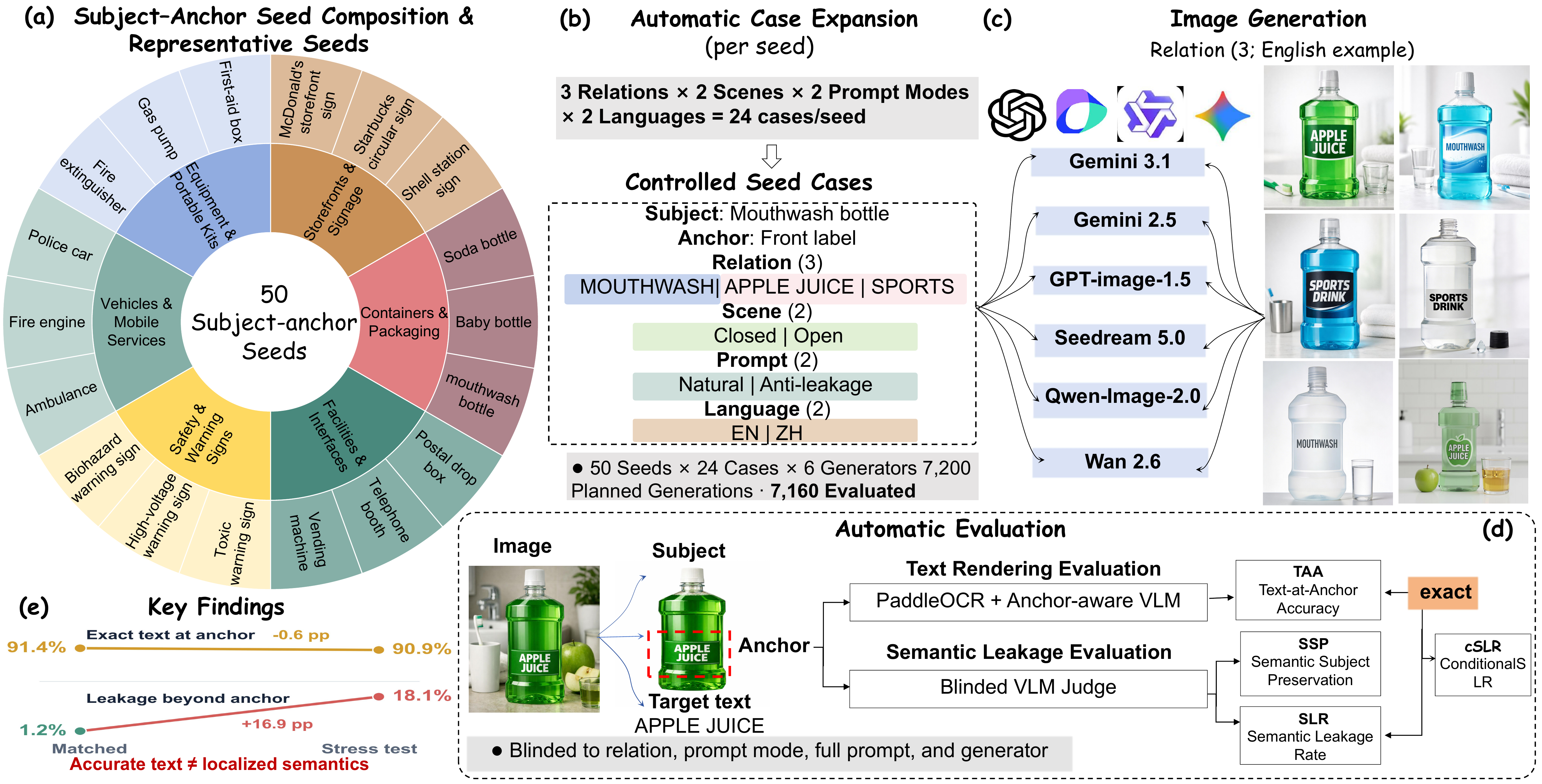}
    \caption{Overview of \bench{} for benchmarking localized semantic control in text-to-image generation.}
    \label{fig:overview}
\end{figure*}

Recent text-to-image models have made substantial progress in image fidelity and instruction following \cite{rombach2022high,betker2023improving}, as well as in explicit text rendering \cite{chen2023textdiffuser,DBLP:conf/iclr/TuoXHGX24}. In practical visual content creation tasks such as storefront signs, device nameplates, and product packaging, target text is typically confined to a designated text-bearing region of a particular subject. These tasks require models not only to render the target text accurately within the designated region, but also to keep its semantic influence strictly local without altering the subject identity or surrounding scene conditions independently specified in the prompt. Such fine-grained local controllability is important for producing multiple variants of advertising and design assets, generating controlled synthetic data \cite{zhu2023conditional}, and creating personalized visual content \cite{ruiz2023dreambooth}. We refer to this requirement as an explicit localized-text contract and define its violation as \textit{target-text-associated semantic leakage}, in which target-text semantics are expressed through non-textual visual content beyond the designated anchor.

However, adhering to this contract is non-trivial. Target text contains both a visual form that must be rendered within the designated region and semantic information that may influence image generation. When the semantics of the target text are inconsistent with the established visual context, including the subject identity, function, and surrounding scene, the model may not treat the text solely as a local rendering constraint, and may alter the subject or scene to make the target text appear more plausible within the overall image. Fig.~\ref{fig:motivation} illustrates this problem in three localized text-rendering scenarios. This raises an important question that remains underexplored: under an explicit localized-text contract, when target text is rendered correctly, can its semantic influence remain confined to the designated region?

Prior studies have identified related forms of semantic interference in generative models. Visual text generation methods may produce the visual concept denoted by a word instead of correctly rendering its glyphs \cite{zhang2026freetext}, while multi-entity generation may transfer semantic attributes from one visual entity to another \cite{ventura2026deleaker}. 
Neither line of work examines the setting studied here, in which the target text is rendered correctly at the designated anchor, yet its semantics propagate beyond that anchor to alter the text-bearing subject or surrounding non-textual content.

Despite the practical importance of semantic containment, existing evaluations are not designed to measure it directly. Benchmarks and methods for text-rich image generation primarily assess whether text is readable, correctly spelled, properly placed, and robustly rendered under complex prompts \cite{DBLP:conf/emnlp/ZhangWLTCTHW25}. Related evaluations of subject preservation and conditional control focus on identity retention or local controllability \cite{li2023blip}. These evaluation settings are important, but they do not explicitly separate text rendering accuracy from semantic containment. Consequently, a model may achieve high text-rendering accuracy even when the rendered text alters the subject or surrounding non-textual scene.

To fill this gap, we present \bench{} (\textbf{Text-to-Image Localized Semantic Control Benchmark}), a controlled diagnostic benchmark designed to stress-test whether text-to-image models can confine target-text semantics to a designated text anchor. \bench{} selects subjects with stable visual identities and default functions, assigns a natural text anchor to each subject, and pairs every seed with one aligned and two conflicting target texts. It further varies scene openness, prompt mode, and language to enable controlled analysis of semantic leakage. The overall construction, generation, and evaluation pipeline of \bench{} is illustrated in Fig.~\ref{fig:overview}. The benchmark comprises 50 seed subjects and 1,200 prompt cases per model, yielding 7,200 planned generations across six models, of which 7,160 are available for evaluation. 
\bench{} measures a model's ability to follow a specific instruction in which the user has localized the text and required the subject and non-textual context to be preserved. This contrasts with open‑ended creative generation, where globally adapting the scene to accommodate new text is considered a reasonable behavior.

We further introduce a dual-branch evaluation protocol that separates anchor-text correctness from semantic containment. The text-rendering branch combines character-level OCR evidence with anchor-aware VLM verification. The semantic-evaluation branch uses a blinded VLM judge to assess subject identity preservation and target-text-associated non-textual evidence outside the designated anchor. The protocol reports Text-at-Anchor Accuracy (TAA), Semantic Subject Preservation (SSP), Semantic Leakage Rate (SLR), and  Conditional Semantic Leakage Rate (cSLR). A stratified human validation study on 420 images shows high agreement between the automatic protocol and adjudicated human annotations.

Experiments on six text-to-image models reveal a clear boundary of localized semantic control. When the target text matches the predefined subject context, models generally preserve both the localized rendering and the surrounding visual semantics. Under the stress-test conditions, however, SLR increases from 1.2\% to 18.1\% and cSLR from 1.3\% to 18.2\%, whereas TAA remains nearly unchanged at 91.4\% and 90.9\%. These results show that a model may follow the local text-rendering instruction while failing to isolate the semantic influence of the rendered text.

\noindent\textbf{Contributions}. In summary, we make the following contributions:
\begin{itemize}[leftmargin=1.5em]
    \item We introduce \bench{}, a controlled benchmark for evaluating localized semantic control under an explicit localized-text contract. Cross-domain target texts serve as diagnostic stress probes for revealing and quantifying target-text-associated semantic leakage. It comprises 50 seed subjects, 1,200 prompt cases per model, and 7,200 planned generations across six text-to-image models, with a factorized design that enables controlled comparison.

    \item We establish a dual-branch evaluation protocol that combines OCR--VLM text verification, blinded VLM-based semantic assessment, and human validation to separately measure anchor-text correctness and semantic containment.
    \item We benchmark six mainstream text-to-image models and demonstrate that stress-test cases expose substantial semantic leakage while having only a limited effect on text-rendering accuracy. We further analyze the effects of prompt mode, scene openness, and language.
\end{itemize}

\section{Background and Related Work}

\noindent\textbf{Visual Text Generation and Evaluation.}
Visual text generation methods improve local textual fidelity through character-aware representations, glyph guidance, and spatial control \cite{liu2023character,chen2023textdiffuser,ma2023glyphdraw,yang2023glyphcontrol}. Subsequent work extends these capabilities to multilingual generation, layout planning, glyph-aligned encoding, text inpainting, and less constrained scenes \cite{zhang2025artist,chen2024textdiffuser,liu2024glyph,zhao2024udifftext,zhu2024visual}. TextCrafter further targets multi-region text rendering in complex scenes \cite{DBLP:journals/corr/abs-2503-23461}. Corresponding benchmarks evaluate readability, spelling, layout, prompt complexity, and rendering robustness \cite{fallah2025textinvision,DBLP:conf/emnlp/ZhangWLTCTHW25}. However, these studies primarily assess whether the requested string is rendered correctly, rather than whether its semantic influence remains confined to the designated text anchor.

\noindent\textbf{Subject Preservation, Compositionality, and Semantic Leakage.}
Subject-driven generation methods preserve identity through learned representations, subject-specific adaptation, and image-based conditioning \cite{ruiz2023dreambooth,li2023blip,ye2023ip}. RefVNLI further evaluates subject preservation together with textual alignment \cite{slobodkin2025refvnli}. Localized generation methods improve region-specific control through explicit spatial conditions and contextual constraints \cite{zhang2023adding,li2023gligen,yang2023reco,DBLP:journals/corr/abs-2510-14553}. Related compositional approaches address entity omission and prompt-structure misalignment \cite{chefer2023attend,feng2022training,rassin2023linguistic}, as well as incorrect attribute binding and unintended concept interactions \cite{li2023dividebind,meral2024conform,kim2025textembedding}. T2I-CompBench and ConceptMix evaluate these failures across attributes, relations, and multiple concepts \cite{huang2023t2icompbench,wu2024conceptmix}. FreeText examines interference between glyph generation and word-associated visual concepts, while DeLeaker studies semantic transfer between visual entities \cite{zhang2026freetext,ventura2026deleaker}. ALE-Bench evaluates attribute leakage in image editing, whereas SemVarBench measures responses to controlled semantic variations \cite{mun2025attributeleakage,zhu2025semvar}. In contrast, \bench{} isolates cases in which the requested text is correctly rendered at a designated anchor, yet its semantics alter the text-bearing subject or surrounding non-textual content.

\noindent\textbf{Semantic Evaluation of Generated Images.}
Automatic evaluation has progressed from global image--text similarity toward structured semantic verification \cite{hessel2021clipscore}. TIFA and GenEval assess prompt fidelity through question answering or object-level checks \cite{hu2023tifa,ghosh2023geneval}, while HEIM, Gecko, and VQAScore broaden evaluation dimensions and compositional assessment \cite{lee2023heim,wiles2024gecko,li2024evaluating}. These methods primarily measure general prompt--image alignment rather than the containment of rendered-text semantics. Our protocol therefore evaluates anchor-text accuracy, subject preservation, and non-textual semantic leakage separately.

\section{Benchmark Construction}

\bench{} is a controlled diagnostic benchmark designed to stress-test whether text-to-image models can confine the semantics of target text to a designated text anchor. Its factorized design supports separate analysis of local text rendering, subject identity preservation, and target-text-associated changes to non-textual visual content under comparable generation conditions.

\subsection{Problem Definition} 

Each benchmark case specifies a text-bearing subject, its predefined identity and default context, a designated text anchor, and a target text. The subject definition is fixed before target-text assignment and serves as the reference for interpreting the generated image.
The benchmark operationalizes an explicit localized-text contract: the target string is to be rendered within the anchor, and the subject identity and non-textual context are to be preserved.
We define \textit{target-text-associated semantic leakage} as the occurrence of visible non-textual cues outside the designated anchor that are semantically linked to the target text but cannot be naturally attributed to the predefined subject context.

\begin{figure}[t]
    \centering
    \includegraphics[width=\columnwidth]{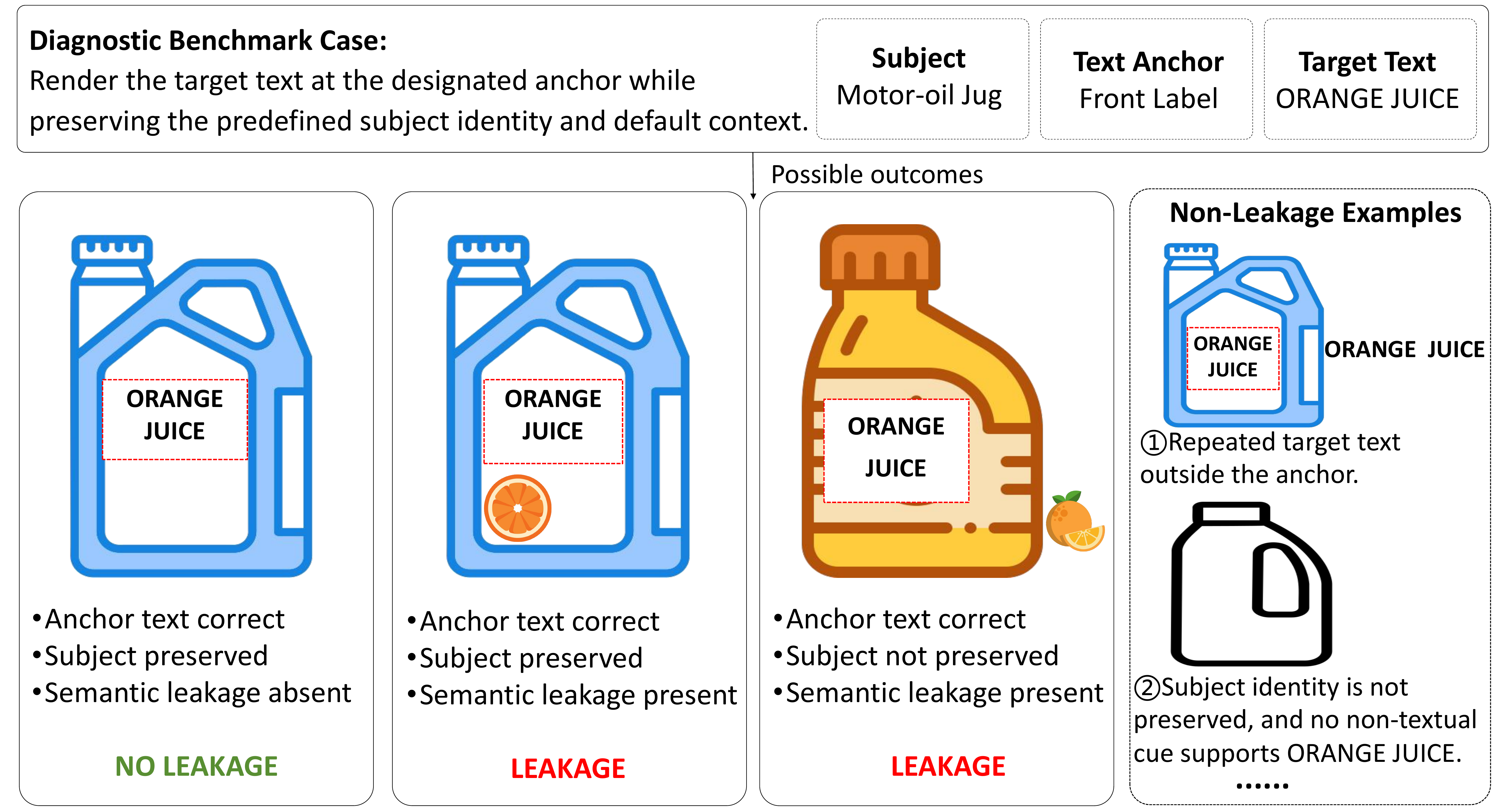}
    \caption{Conceptual boundary of target-text-associated semantic leakage. }
    \label{fig:conceptual_boundary}
\end{figure}

This definition is deliberately normative about the contract but neutral about generation in general. A model that globally adapts a scene to accommodate new text under an underspecified prompt is not wrong; \bench{} tests the narrower, practically important ability to respect an explicit localization instruction.
Fig.~\ref{fig:conceptual_boundary} highlights two important distinctions. First, subject preservation and semantic leakage are separate properties: target-text-associated cues may appear on an otherwise recognizable subject or in its surrounding scene. Second, not every generation error constitutes semantic leakage. Extra or repeated text, target-related content confined to the designated anchor, cues expected in the default subject context, and subject degradation unrelated to the target text are not counted as semantic leakage. Text rendering, subject preservation, and semantic leakage are operationalized separately in Section~\ref{sec:evaluation_protocol}.

\begin{figure*}[t]
\centering
\includegraphics[width=0.92\textwidth]{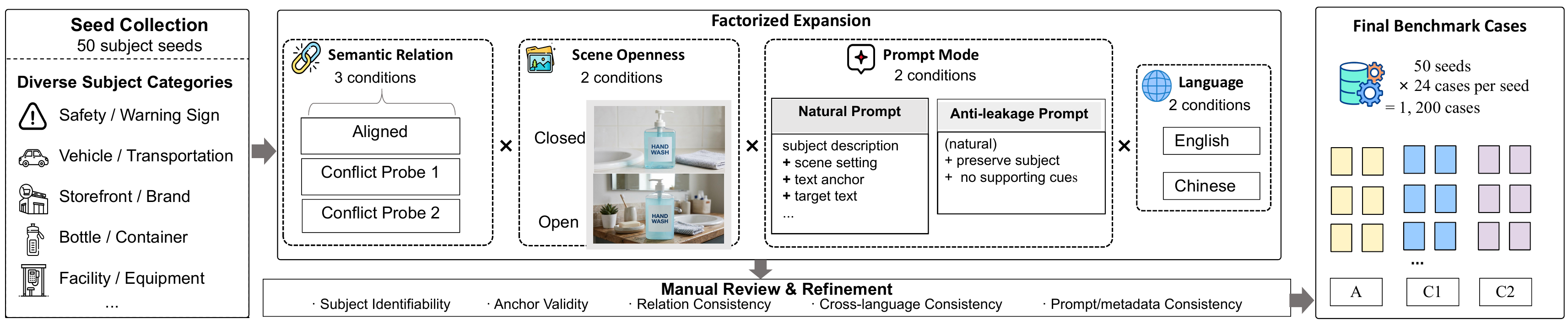}
\caption{Factorized case instantiation and quality control in \bench{}.}
\label{fig:case_instantiation}
\end{figure*}

\subsection{Seed and Text-Anchor Design}

Seed selection emphasizes both the observability of semantic leakage and the reliability of its evaluation. Each seed must have a stable visual identity, a native text-bearing region, and a clear default semantic or functional interpretation. These properties enable us to distinguish subject-identity failures from ordinary generation variations and to determine whether non-textual changes express the target-text semantics beyond what is naturally expected from the default subject context.

To reduce ambiguity, we exclude abstract scene categories and place-level descriptions that do not specify a concrete visual subject or a clearly identifiable text-bearing region. Instead, each seed defines a concrete, visually recognizable subject and a native text anchor, such as a storefront facade with a mounted signboard, a service vehicle with a body-side marking region, a product container with a front label, or a warning sign with a predefined text panel. The anchor specifies the intended location of the target string; rendering the correct string elsewhere therefore does not count as successful anchor-text rendering. The subject definition establishes the reference identity and default context, whereas the text anchor defines the local boundary for evaluating text rendering and semantic leakage. The 50 seeds cover six descriptive subject categories. Representative seeds are shown in Fig.~\ref{fig:overview}(a), while the complete seed inventory is provided in Table~S1 of the supplementary material.

\subsection{Controlled Factors}

\bench{} varies four controlled factors to test complementary hypotheses about target-text-associated semantic leakage. Semantic relation tests whether localized semantic control remains reliable when the target text is semantically incongruent with the predefined subject context. Prompt mode tests whether explicit semantic-boundary constraints suppress leakage, while scene openness tests whether richer non-textual context provides more potential carriers for target-text semantics. Language tests whether the phenomenon is specific to a particular language or writing system. The subject and text anchor remain fixed across matched cases, supporting controlled comparisons of these factors. Their operational definitions are summarized in Table~\ref{tab:controlled_condition_spec}.

\begin{table}[htbp]
\centering
\small
\caption{Operational specification of the controlled factors in \bench{}.}
\label{tab:controlled_condition_spec}
\resizebox{\linewidth}{!}{
\begin{tabular}{
>{\centering\arraybackslash}m{2.2cm}
>{\raggedright\arraybackslash}m{1.5cm}
>{\raggedright\arraybackslash}m{8.8cm}
}
\toprule[1.2pt]
\textbf{Factor}
& \textbf{Condition}
& \textbf{Operational specification} \\
\midrule[1.2pt]

\multirow[c]{2}{=}{Semantic relation}
& aligned
& The target text is semantically compatible with the predefined subject identity and default function. \\
\cmidrule{2-3}
& conflicting
& The target text conflicts with the predefined subject interpretation. Each seed includes two distinct conflict probes. \\

\midrule

\multirow[c]{2}{=}{Prompt mode}
& natural
& A controlled base prompt specifying the subject, scene openness, text anchor, target text, readability, and native placement. \\
\cmidrule{2-3}
& anti\_leakage
& The same base prompt with additional constraints against semantic propagation, subject or scene reinterpretation, supporting visual content, and scene reconstruction. \\

\midrule

\multirow[c]{2}{=}{Scene openness}
& closed
& A subject-centered composition with 1--3 surrounding objects, low background density, and a mid-shot view. \\
\cmidrule{2-3}
& open
& A context-rich composition with 4--6 surrounding objects, medium background density, and a mid-to-far-shot view, while keeping the subject identifiable. \\

\midrule

\multirow[c]{2}{=}{Language}
& English
& The subject description, text anchor, target text, and prompt template are instantiated in English. \\
\cmidrule{2-3}
& Chinese
& The corresponding components are instantiated in Chinese while preserving the same subject--relation design. \\

\bottomrule[1.2pt]
\end{tabular}
}
\end{table}

\subsubsection{Semantic Relation}

Each seed is paired with one aligned target text and two conflicting target texts. The aligned text is compatible with the predefined subject identity and default function, providing a reference condition for ordinary text--subject consistency. By contrast, the conflicting texts are semantically incompatible with the default subject interpretation and short enough to fit within the designated anchor. Each text conveys a clear and specific meaning, reducing ambiguity in the subsequent leakage assessment. They cover functional, category, service-domain, product-domain, and safety-context mismatches. These mismatch categories are used only to diversify the conflict probes and reduce dependence on any single target text.

\subsubsection{Prompt Mode}

The natural condition uses a controlled base prompt that specifies the subject, scene openness, text anchor, target text, readability, and native-placement requirements. The anti\_leakage condition retains the same base prompt while adding constraints against semantic propagation from the target text, subject or scene reinterpretation, supporting non-textual content, and scene reconstruction. All remaining prompt content is unchanged, allowing the effect of semantic-boundary constraints to be isolated.

\subsubsection{Scene Openness}

Scene openness controls the amount of non-textual context available around the subject.The closed condition specifies a subject-centered mid-shot composition with 1--3 surrounding objects and low background density, whereas the open condition uses a mid-to-far-shot composition with 4--6 surrounding objects and medium background density while keeping the main subject identifiable. These settings provide a reproducible contrast in contextual richness while avoiding both isolated object views and excessively complex scenes. Comparing the two conditions tests whether richer surroundings provide more opportunities for target-text-associated objects or contextual cues to appear outside the designated text anchor.

\subsubsection{Language}

Each case is instantiated in English and Chinese using language-specific subject descriptions, text anchors, target texts, and prompt templates. The subject identity, semantic relation, scene openness, and prompt mode are preserved across the paired conditions, enabling a controlled comparison between the two distinct language conditions.

Together, these factors define the controlled condition space used for factorized case instantiation and subsequent analysis.

\subsection{Case Instantiation and Quality Control}

\bench{} constructs benchmark cases through a full factorial expansion of the controlled factors. Each seed specifies a subject description, a text-anchor specification, one aligned target text, and two conflicting target texts. For each of the 50 seeds, the three target-text instances are combined with two scene-openness settings, two prompt modes, and two languages. This process yields 1,200 prompt instances, all of which are applied to each evaluated model. The overall case-instantiation process is illustrated in Fig.~\ref{fig:case_instantiation}.

We conduct consistency checks at the subject, anchor, target-text, and cross-language levels. Subject descriptions are reviewed to ensure that they refer to concrete physical entities with recognizable identities. Text anchors are checked for naturalness, visual specificity, and compatibility with the corresponding subjects. Aligned target texts are verified to be consistent with the default subject interpretation, whereas conflicting target texts are checked to ensure that they are semantically incompatible with that interpretation while remaining concrete and suitable for the designated anchor. The English and Chinese versions are reviewed to preserve the same underlying subject, anchor, and semantic-relation design.

\begin{figure*}[t]
    \centering
    \includegraphics[width=0.93\textwidth]{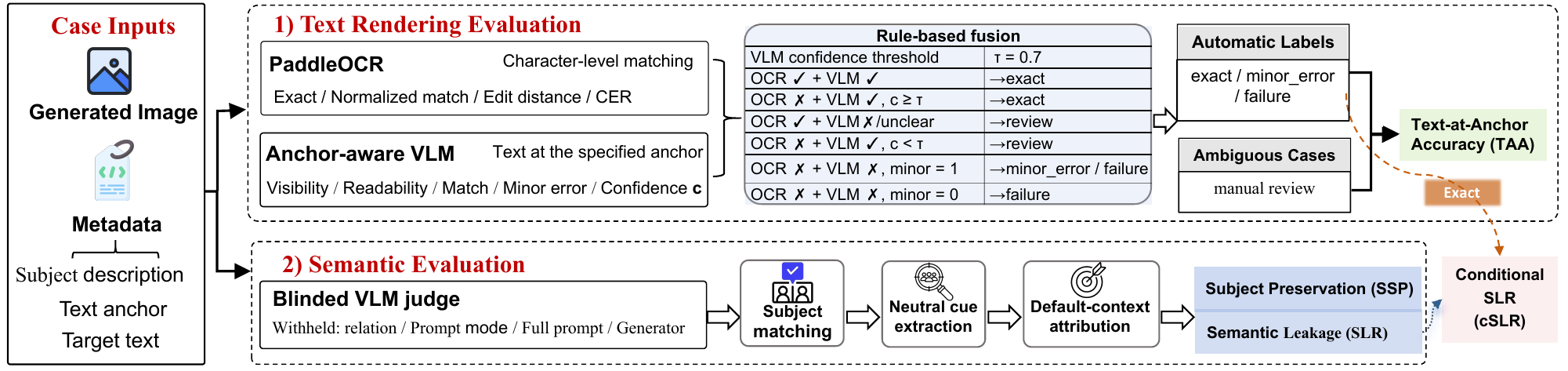}
   \caption{Overview of the automatic evaluation protocol.}
    \label{fig:evaluation_protocol}
\end{figure*}
\section{Evaluation}

\subsection{Experimental Setup}

We evaluate six text-to-image models: Doubao Seedream 5.0, Gemini 3.1 Flash Image Preview, Gemini 2.5 Flash Image, GPT-image-1.5, Wan2.6 Image, and Qwen-Image-2.0. For brevity, we refer to them as Seedream 5.0, Gemini 3.1, GPT-image-1.5, Wan2.6, Qwen-2.0, and Gemini 2.5, respectively. Each model is assigned the same 1,200 benchmark cases, yielding 7,200 planned generations. The final evaluation includes 7,160 available images: 1,176 from GPT-image-1.5, 1,184 from Qwen-Image-2.0, and 1,200 from each of the other four models. The unavailable outputs all involve the \textit{Disney storefront} seed and were blocked by provider-side content-safety filters.

All images are evaluated using a shared protocol and judge configuration. The protocol combines OCR evidence, VLM-based anchor-text verification, and VLM-based semantic assessment. Only OCR--VLM disagreements that cannot be resolved by the predefined fusion rules are referred for manual review. Results are analyzed across the four controlled factors defined in the preceding section.

Separately, we validate the automatic protocol on a stratified subset of 420 images, with 70 images sampled from each model. Within each model, the subset contains 35 automatically predicted SLR-positive and 35 SLR-negative cases while maintaining coverage of semantic relation, scene openness, prompt mode, and language. Two authors with experience in text-to-image evaluation independently annotate TAA, SSP, and SLR without access to the automatic predictions. Disagreements are adjudicated by a third author with experience in multimodal evaluation, and the resulting labels serve as references for human validation.

\subsection{Evaluation Protocol}
\label{sec:evaluation_protocol}

As illustrated in Fig.~\ref{fig:evaluation_protocol}, the protocol comprises two evaluation branches that are applied independently to each image. The text-rendering branch combines character-level OCR evidence with anchor-aware VLM verification to determine whether the target text is correctly rendered at the designated anchor. Independently of the text-rendering outcome, the semantic branch evaluates every image for subject identity preservation and target-text-associated semantic leakage. An uncertain outcome is assigned when the visible evidence is insufficient to support a reliable judgment. The conditional leakage measure is subsequently computed over samples with exact anchor-text rendering and a valid semantic-leakage judgment.

We use \texttt{gemini-3.1-pro-preview} with the temperature set to 0 for both text-rendering evaluation and semantic evaluation. The two tasks use separate fixed prompts and JSON output schemas. The VLM version, prompts, confidence thresholds, OCR--VLM fusion rules, and output-parsing procedures remain fixed across all images and evaluated generators.

\subsubsection{Text-Rendering Evaluation}

The text-rendering branch determines whether the complete target string is correctly rendered within the designated native text-bearing region. We first apply PaddleOCR using its Chinese recognition configuration, which supports both Chinese and Latin characters, with the angle classifier enabled. OCR detections with confidence below 0.5 are discarded to suppress spurious responses to background patterns, decorative elements, and unreadable text-like shapes. The retained detections are compared with the target string using exact match, normalized match, edit distance, and character error rate. Exact and normalized matches are used in the fusion decision, whereas edit distance and character error rate serve as diagnostic evidence. OCR provides character-level evidence but cannot reliably determine whether a detected string appears at the designated anchor. It may also fragment multi-line text or incorrectly recognize malformed glyphs as valid characters.

To verify both text content and placement, an anchor-aware VLM examines the designated text region and reports the detected string, visibility, readability, placement validity, character-level deviations, ambiguity, and confidence $c_i\in[0,1]$. Let $O_i=1$ denote an exact or normalized OCR match and $V_i=1$ denote the corresponding VLM match, with a confidence threshold of $\tau=0.7$. A sample is provisionally labeled as \textit{exact} if either $O_i=V_i=1$ or $O_i=0$, $V_i=1$, and $c_i\geq\tau$. The second rule recovers likely OCR false negatives caused by missed or fragmented detections when the target text is nevertheless visibly correct at the designated anchor.

All OCR--VLM disagreements that remain unresolved by the predefined fusion rules, including OCR matches rejected by the VLM and VLM matches below the confidence threshold, are referred for manual review. Following fusion and review, a non-exact sample with a small but visible character-level deviation is labeled as \textit{minor\_error}, whereas other visibly incorrect renderings are labeled as \textit{failure}. The label \textit{ambiguous} is reserved for cases in which the available visual evidence remains insufficient after review. Only samples assigned the final label \textit{exact} are counted as successful anchor-text renderings.

\subsubsection{Semantic Evaluation}

The semantic-evaluation branch evaluates whether the image contains target-text-associated non-textual visual evidence outside the designated anchor. The semantic judge receives only the generated image, subject description, text anchor, and target text. The semantic-relation label, prompt mode, complete generation prompt, and generator identity are withheld to prevent the controlled condition or model identity from influencing the judgment.

Rather than requesting a direct leakage decision, the prompt decomposes semantic evaluation into three evidence-grounded steps. The judge first locates the visual entity corresponding to the predefined subject and assesses whether its identity is preserved. It then describes candidate non-textual cues outside the designated anchor in neutral terms. For each cue, the judge determines whether it supports the target-text semantics and whether it would be naturally expected from the default subject context in the absence of the target text. A cue is treated as leakage evidence only when it supports the target-text semantics but cannot be naturally explained by the default subject context. This procedure produces separate SSP and SLR judgments together with concise evidence grounded in visible image content.

Text-only artifacts are treated separately from semantic leakage. Repeated target text and other readable text outside the designated anchor, including text on external signs, labels, and captions, do not constitute SLR by themselves. Likewise, generic blur, structural deformation, or loss of subject identity does not by itself indicate semantic leakage. Such a change contributes to SLR only when it provides visible non-textual evidence that supports the target-text semantics beyond the default subject context.

\subsubsection{Evaluation Metrics}
We report four complementary metrics.  
\begin{itemize}[leftmargin=15pt]
\item \textbf{Text-at-Anchor Accuracy (TAA)} measures the proportion of available images in which the complete target string is correctly rendered at the designated anchor.
\item \textbf{Semantic Subject Preservation (SSP)} measures the proportion of valid subject-preservation judgments in which the predefined subject identity remains recognizable.
\item \textbf{Semantic Leakage Rate (SLR)} measures the proportion of valid leakage judgments in which non-textual visual content outside the designated anchor expresses target-text semantics beyond what would naturally be expected from the default subject context.
\item \textbf{Conditional Semantic Leakage Rate (cSLR)} measures the same leakage outcome specifically among samples with exact anchor-text rendering.
\end{itemize}

Let $\mathcal{D}$ denote the set of all available images, and let $\mathcal{V}_{S}$ and $\mathcal{V}_{L}$ denote the subsets for which valid binary SSP and SLR judgments are available, respectively. For each image $i\in\mathcal{D}$, let $q_i$ denote the final text-rendering label. For each $i\in\mathcal{V}_{S}$, let $s_i\in\{0,1\}$ denote the subject-preservation judgment, where $s_i=1$ indicates that the predefined subject identity is preserved. For each $i\in\mathcal{V}_{L}$, let $l_i\in\{0,1\}$ denote the leakage judgment, where $l_i=1$ indicates the presence of semantic leakage. Let $\mathbb{I}[\cdot]$ denote the indicator function. We define

\begin{align}
\mathrm{TAA}
&=
\frac{100}{|\mathcal{D}|}
\sum_{i\in\mathcal{D}}
\mathbb{I}[q_i=\textit{exact}], \\
\mathrm{SSP}
&=
\frac{100}{|\mathcal{V}_{S}|}
\sum_{i\in\mathcal{V}_{S}}
\mathbb{I}[s_i=1], \\
\mathrm{SLR}
&=
\frac{100}{|\mathcal{V}_{L}|}
\sum_{i\in\mathcal{V}_{L}}
\mathbb{I}[l_i=1].
\end{align}

Because SLR is computed over both exact and non-exact text renderings, it does not isolate whether leakage persists after successful local text rendering. We therefore define

\begin{equation}
\mathcal{C}_{L}
=
\left\{
i\in\mathcal{V}_{L}
\mid
q_i=\textit{exact}
\right\},
\end{equation}
where $\mathcal{C}_{L}$ contains samples with exact anchor-text rendering and a valid leakage judgment. Conditional Semantic Leakage Rate is defined as

\begin{equation}
\mathrm{cSLR}
=
\frac{100}{|\mathcal{C}_{L}|}
\sum_{i\in\mathcal{C}_{L}}
\mathbb{I}[l_i=1].
\end{equation}

cSLR quantifies semantic leakage conditional on successful anchor-text rendering and therefore directly assesses whether target-text semantics remain locally contained after the target string has been rendered exactly.

All metrics are reported as percentages. Higher TAA and SSP indicate better performance, whereas lower SLR and cSLR indicate stronger semantic containment. An uncertain SSP judgment is excluded only from the denominator of SSP, whereas an uncertain SLR judgment is excluded from the denominators of both SLR and cSLR. cSLR further requires an \textit{exact} text-rendering label. To quantify the semantic-relation contrast, we additionally report

\begin{align}
\Delta\mathrm{SLR}
&=
\mathrm{SLR}_{\mathrm{cf}}
-
\mathrm{SLR}_{\mathrm{al}},
\end{align}
where $\mathrm{al}$ denotes the aligned condition and $\mathrm{cf}$ pools the two conflict probes defined for each seed.

\begin{table}[t]
\centering
\small
\setlength{\tabcolsep}{4.2pt}
\renewcommand{\arraystretch}{1.1}
\caption{Overall and relation-specific performance on \bench{} (\%).}
\label{tab:overall_results}
\begin{tabular}{@{}lrrrrrrrr@{}}
\toprule
& &
\multicolumn{2}{c}{\textbf{Overall}}
& \multicolumn{2}{c}{\textbf{SLR$\downarrow$}}
& \multicolumn{2}{c}{\textbf{cSLR$\downarrow$}}
& \textbf{$\Delta$SLR$\downarrow$} \\
\cmidrule(lr){3-4}
\cmidrule(lr){5-6}
\cmidrule(lr){7-8}
\textbf{Model}
& \textbf{$N$}
& \textbf{TAA}
& \textbf{SSP}
& \textbf{Al.}
& \textbf{Cf.}
& \textbf{Al.}
& \textbf{Cf.}
& \textbf{pp} \\
\midrule
Seedream 5.0
& 1200 & 97.1 & 96.4
& 1.8 & 21.5 & 1.8 & 22.0 & 19.7 \\

Gemini 3.1
& 1200 & \textbf{99.0} & \textbf{98.8}
& \textbf{0.5} & \textbf{13.0}
& \textbf{0.5} & \textbf{13.1}
& \textbf{12.5} \\

GPT-1.5
& 1176 & 98.3 & 96.9
& \textbf{0.5} & 14.0
& \textbf{0.5} & 14.2
& 13.5 \\

Wan2.6
& 1200 & 94.2 & 91.7
& 2.0 & 22.4 & 2.1 & 23.4 & 20.4 \\

Qwen-2.0
& 1184 & 98.8 & 96.5
& 1.5 & 15.8 & 1.6 & 15.9 & 14.3 \\

Gemini 2.5
& 1200 & 59.2 & 96.7
& 1.0 & 21.9 & 1.3 & 22.7 & 20.9 \\

\midrule
All
& 7160 & 91.0 & 96.2
& 1.2 & 18.1 & 1.3 & 18.2 & 16.9 \\
\bottomrule
\end{tabular}

\vspace{1pt}
\parbox{\columnwidth}{
\scriptsize\raggedright
\textit{Note.}
$N$ denotes the number of available images. Metric-specific valid denominators are reported in the supplementary material. The final row reports image-level micro-averages.
}
\end{table}

\section{Experimental Results}
\subsection{Overall Results}
\autoref{tab:overall_results} shows that text-rendering accuracy and semantic control do not necessarily vary together. TAA ranges from 59.2\% to 99.0\%, whereas SSP ranges from 91.7\% to 98.8\%. Under conflicting conditions, aggregate SLR reaches 18.1\%, and restricting the analysis to exact text renderings yields a nearly identical cSLR of 18.2\%. Thus, correctly rendering the requested text does not necessarily indicate effective control over its semantic influence.

The model-level results reveal distinct capability profiles. Gemini 3.1 achieves the strongest overall performance, although its conflicting cSLR remains 13.1\%. Seedream 5.0, by contrast, combines a high TAA of 97.1\% with a conflicting cSLR of 22.0\%, indicating strong glyph rendering but limited containment of text semantics. Wan2.6 records both the lowest SSP and the highest conflicting cSLR, suggesting that semantic conflict affects both subject identity and surrounding visual content. Gemini 2.5 exhibits limitations in both text rendering and semantic containment, although its cSLR is computed from a smaller exact-rendering subset. These differences are consistent with varying degrees of coupling between local text rendering and global scene synthesis.

\begin{figure*}[t]
\centering
\includegraphics[width=0.91\textwidth]{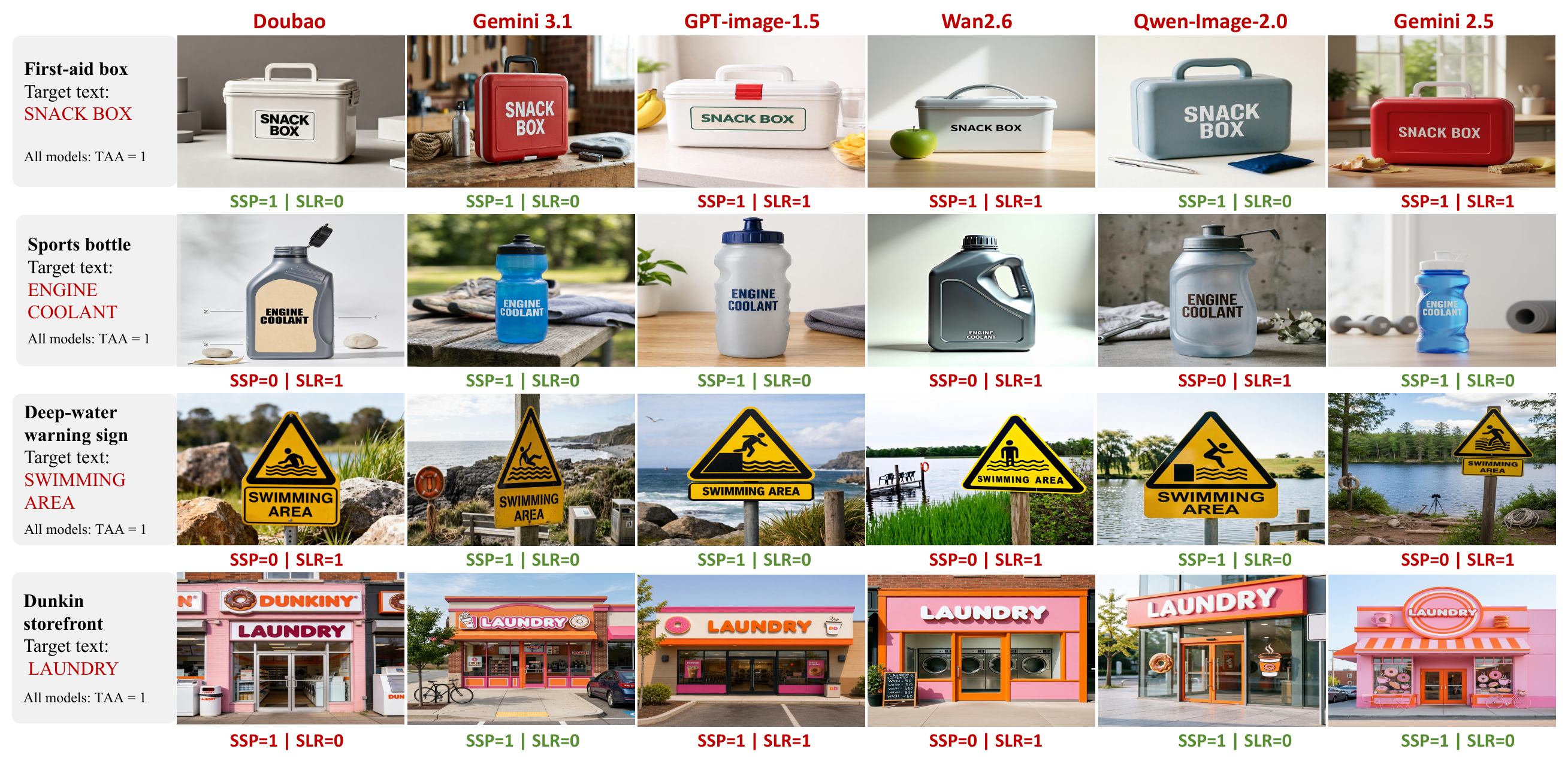}
\caption{Cross-model comparison under matched benchmark conditions. All outputs achieve exact anchor-text rendering but differ in subject preservation and semantic leakage.}
\label{fig:cross_model_comparison}
\end{figure*}

Fig.~\ref{fig:cross_model_comparison} illustrates these differences under matched conditions. In the first-aid-box case, several models introduce food-related objects associated with ``SNACK BOX,'' while others preserve the default context. In the sports-bottle case, leakage appears as reinterpretation of the subject as an automotive-fluid container. The warning-sign and storefront cases further show target-text-associated modification of a non-textual pictogram and broader reconstruction of the storefront as a laundry, respectively. These examples demonstrate that the same semantic conflict can produce object insertion, subject reinterpretation, or scene reconstruction depending on the model, even when anchor-text accuracy is held constant.

Of the 7,200 planned generations, 7,160 are available for evaluation. The 40 unavailable outputs are associated with the \textit{Disney storefront} seed and provider-side content-safety filtering. Excluding this seed from all models changes every reported metric by less than 0.5 percentage points, indicating that the main findings are not driven by the missing outputs.

\subsection{Effects of Controlled Factors}

\begin{table}[t]
\centering
\small
\setlength{\tabcolsep}{0.8pt}
\renewcommand{\arraystretch}{1.0}
\caption{Performance under the controlled factors of \bench{} (\%).}
\label{tab:controlled_factor_results}

\begin{tabular*}{\columnwidth}{
@{\extracolsep{\fill}}llrccc@{}}
\toprule
\textbf{Factor}
& \textbf{Condition}
& \textbf{$N$}
& \textbf{TAA$\uparrow$}
& \textbf{SSP$\uparrow$}
& \textbf{SLR$\downarrow$} \\
\midrule

\multirow{2}{*}{Relation}
& $R_{\mathrm{al}}$
& 2384
& 91.4
& 98.7(2381)
& 1.2(2381) \\

& $R_{\mathrm{cf}}$
& 4776
& 90.9
& 94.9(4773)
& 18.1(4773) \\

\midrule

\multirow{2}{*}{Prompt}
& $P_{\mathrm{nat}}$
& 3582
& 91.0
& 95.8(3581)
& 16.6(3581) \\

& $P_{\mathrm{anti}}$
& 3578
& 91.1
& 96.5(3573)
& 8.4(3573) \\

\midrule

\multirow{2}{*}{Scene}
& $S_{\mathrm{cl}}$
& 3580
& 91.1
& 96.2(3577)
& 11.4(3577) \\

& $S_{\mathrm{op}}$
& 3580
& 91.0
& 96.1(3577)
& 13.6(3577) \\

\midrule

\multirow{2}{*}{Language}
& $L_{\mathrm{en}}$
& 3582
& 97.9
& 95.6(3579)
& 14.1(3579) \\

& $L_{\mathrm{zh}}$
& 3578
& 84.2
& 96.8(3575)
& 10.9(3575) \\

\bottomrule
\end{tabular*}

\vspace{1pt}
\parbox{\columnwidth}{
\footnotesize\raggedright
\textit{Note.}
$N$ denotes the number of available images. Values in parentheses are
valid denominators for SSP and SLR. The results are image-level
micro-averages. Relation, prompt, and scene denote semantic relation,
prompt mode, and scene openness, respectively.
}
\end{table}

\autoref{tab:controlled_factor_results} reports image-level micro-averages over the six models. For each controlled factor, images from the remaining conditions are pooled, and each image receives equal weight. Model-level macro-averaging yields nearly identical SLR values and the same overall trends. To account for correlations among cases derived from the same subject, we compute paired confidence intervals using 10,000 bootstrap samples over the 50 subject seeds.

\begin{figure}[t]
    \centering
    \includegraphics[width=0.92\columnwidth]{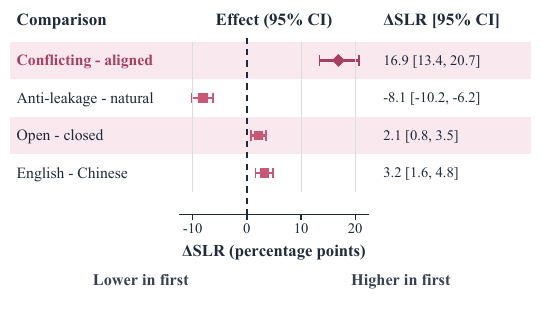}
    \caption{Paired seed-clustered bootstrap estimates of SLR differences between controlled conditions. Points indicate estimated differences, horizontal bars indicate 95\% confidence intervals, and the dashed line denotes no difference.}
    \label{fig:bootstrap_slr_effects}
\end{figure}

As shown in \autoref{fig:bootstrap_slr_effects}, all confidence intervals exclude zero. Semantic relation yields the largest SLR difference, followed by prompt mode, whereas scene openness and language exhibit smaller differences. Overall, leakage varies most strongly with semantic relation, while explicit anti-leakage prompting substantially, but not completely, reduces it.

\paragraph{Semantic relation}
The semantic-relation comparison provides the primary stress test of localized semantic control. Compared with the semantically matched condition $R_{\mathrm{al}}$, the stress-test condition $R_{\mathrm{cf}}$ increases SLR from 1.2\% to 18.1\%. The seed-clustered difference is 16.9 percentage points, with a 95\% confidence interval of $[13.4,20.7]$. By contrast, TAA decreases by only 0.5 percentage points. Moreover, cSLR increases from 1.3\% to 18.2\%, showing that leakage remains prevalent even when the target text is rendered exactly.  Together, these results show that models may successfully follow the local text-rendering instruction while failing to confine the semantic influence of the target text to the designated anchor.

\paragraph{Prompt mode}
Explicit anti-leakage instructions provide substantial but incomplete mitigation. SLR decreases from 16.6\% under $P_{\mathrm{nat}}$ to 8.4\% under $P_{\mathrm{anti}}$, while TAA remains nearly unchanged. The seed-clustered difference is $-8.1$ percentage points, with a 95\% confidence interval of $[-10.2,-6.2]$. The magnitude of this reduction varies substantially across models: GPT-image-1.5 shows a large decrease, whereas Seedream 5.0 changes only marginally. Explicit instructions can therefore reduce semantic leakage, but they do not provide consistent semantic containment across models.

\paragraph{Scene openness}
Open scenes exhibit a higher SLR than closed scenes (13.6\% versus 11.4\%). The seed-clustered difference is 2.1 percentage points, with a 95\% confidence interval of $[0.8,3.5]$. Richer visual contexts may provide more opportunities for target-text-associated non-textual cues to appear outside the designated anchor, although this difference is smaller than those associated with semantic relation and prompt mode.

\paragraph{Language}
The language comparison reveals distinct patterns for text rendering and semantic leakage. Across all six models, English cases exhibit higher SLR than Chinese cases, with aggregate SLR values of 14.1\% and 10.9\%, respectively. The difference ranges from 0.7 percentage points for Gemini 3.1 to 5.2 percentage points for Wan2.6, indicating a consistent association whose magnitude varies across models. In contrast, the large aggregate TAA difference (97.9\% versus 84.2\%) is driven primarily by the lower Chinese text-rendering accuracy of Gemini 2.5 and therefore does not reflect a consistent cross-model language pattern. Overall, the language condition is consistently associated with semantic leakage across the evaluated models, whereas the aggregate difference in text-rendering accuracy is dominated by model-specific capability.

\subsection{Human Validation Results}

\begin{table}[t]
\centering
\scriptsize
\setlength{\tabcolsep}{5pt}
\caption{Human validation on the stratified 420-image subset.}
\label{tab:human_validation}

\textbf{(a) Confusion counts}

\vspace{2pt}
\begin{tabular}{lrrrr}
\toprule
\textbf{Metric}
& \textbf{TP}
& \textbf{FP}
& \textbf{TN}
& \textbf{FN} \\
\midrule
TAA & 338 & 12 & 63  & 7  \\
SSP & 338 & 7  & 61  & 14 \\
SLR & 175 & 35 & 203 & 7  \\
\bottomrule
\end{tabular}

\vspace{6pt}

\textbf{(b) Agreement and classification performance}

\vspace{2pt}
\begin{tabular}{lrrrrr}
\toprule
\textbf{Metric}
& \textbf{Agr. (\%)}
& \textbf{$\kappa$}
& \textbf{Prec. (\%)}
& \textbf{Rec. (\%)}
& \textbf{F1 (\%)} \\
\midrule
TAA & 95.5 & 0.842 & 96.6 & 98.0 & 97.3 \\
SSP & 95.0 & 0.823 & 98.0 & 96.0 & 97.0 \\
SLR & 90.0 & 0.800 & 83.3 & 96.2 & 89.3 \\
\bottomrule
\end{tabular}

\vspace{2pt}
\parbox{\linewidth}{
\footnotesize
\raggedright
\textit{Note.}
Human annotations are treated as reference labels and automatic judgments as predictions.
Agr., Prec., and Rec. denote agreement, precision, and recall.
}
\end{table}
\begin{figure*}[t]
\centering
\includegraphics[width=0.88\textwidth]{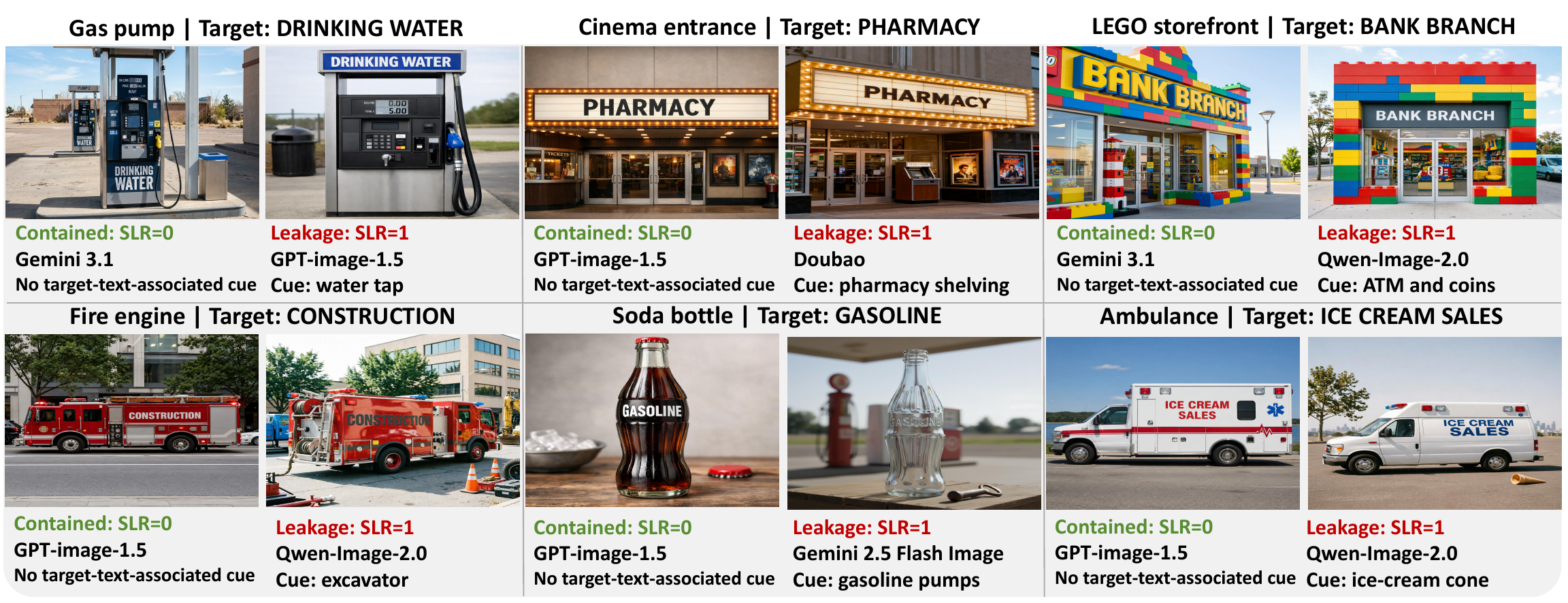}
\caption{Matched comparisons of semantic containment and leakage. Each pair shares the same subject, target text, text anchor, and controlled conditions, although the outputs may come from different models. All outputs satisfy $\mathrm{TAA}=\mathrm{SSP}=1$; only leakage cases satisfy $\mathrm{SLR}=1$.}
\label{fig:contained_leakage_comparison}
\end{figure*}
As shown in \autoref{tab:human_validation}, the automatic protocol exhibits high agreement with the adjudicated human annotations across all three dimensions. TAA and SSP achieve agreement rates of 95.5\% and 95.0\%, with Cohen's $\kappa$ values of 0.842 and 0.823, respectively. Their precision, recall, and F1 scores all exceed 96\%, indicating reliable assessment of anchor-text correctness and subject preservation.

SLR remains more challenging because its assessment requires distinguishing target-text-associated non-textual cues from the surrounding context that is naturally compatible with the predefined subject. Nevertheless, the automatic SLR judge achieves 90.0\% agreement, a Cohen's $\kappa$ of 0.800, and an F1 score of 89.3\%. Its recall of 96.2\% shows that most human-confirmed leakage cases are detected, whereas its lower precision of 83.3\% reflects a greater tendency toward false-positive leakage judgments. Inspection of these false positives suggests that the main remaining difficulty is the over-attribution of ambiguous or otherwise natural contextual cues to the target-text semantics. Overall, the automatic evaluation shows high agreement with human annotations, while further improvement is needed to distinguish target-text-associated semantic evidence from visual cues that are natural to the predefined subject context.

\subsection{Qualitative Observations}

Fig.~\ref{fig:contained_leakage_comparison} compares semantic containment and leakage under matched benchmark conditions. Leakage is indicated by target-text-associated non-textual cues outside the designated anchor, including a water tap, pharmacy shelving, an ATM and coins, an excavator, gasoline pumps, and an ice-cream cone. By contrast, the corresponding contained outputs render the same target text without introducing such cues. These comparisons clarify the SLR decision boundary: Leakage is identified only when the semantics of the target text are reflected in visible non-textual content outside the designated anchor and cannot be naturally explained by the default subject context.

\section{Discussion}

\bench{} provides a controlled testbed for comparing how effectively text-to-image models confine target-text semantics to a designated region while preserving the subject and scene. It can support cross-model comparison, regression testing across model updates, and the evaluation of prompt or training-based strategies for reducing leakage, with applications to advertising and design variants, controlled synthetic data, and personalized visual content. However, the current benchmark focuses on subjects with predefined text anchors and stable visual priors, relatively short English and Chinese target strings. Future work could extend the evaluation to longer target strings, additional languages, less structured subjects, and image and video editing settings. 

\section{Conclusion}

We introduced \bench{}, a controlled diagnostic benchmark for evaluating localized-text semantic containment in T2I generation. By separately measuring anchor-text accuracy, subject preservation, and semantic leakage under an explicit localized-text contract, \bench{} reveals that correct text rendering, by itself, does not guarantee local semantic containment. 
Across six models, semantically matched cases generally preserve localized control, whereas the stress-test cases expose substantial semantic leakage despite only minor changes in text-rendering accuracy. Explicit anti-leakage prompting offers partial, model-dependent mitigation.
These findings establish semantic containment as an important and complementary dimension of visual-text generation and evaluation, and provide a reproducible protocol for measuring it in current and future multimedia content-generation systems.

\bibliographystyle{IEEEtran}
\bibliography{sample-base}

\end{document}


\title{Supplementary Material for ``T2LSC-Bench: Benchmarking Localized Semantic Control in Text-to-Image Generation''}

\author{Yan Wang, Xinyi Hou, Weiguo Lin, Junjun Si, and Siwei Ma}

\maketitle

\section{Complete Benchmark Seed Inventory}

Table~\ref{tab:full_seed_composition} lists the complete set of 50 subject seeds and their predefined text anchors. The seed definitions are fixed before assigning aligned and conflicting target texts.

\begin{table*}[t]
\centering
\small
\setlength{\tabcolsep}{5pt}
\renewcommand{\arraystretch}{1.12}
\caption{Complete inventory of the 50 subject--anchor seeds in \bench{}.}
\label{tab:full_seed_composition}
\begin{tabularx}{\textwidth}{
>{\raggedright\arraybackslash}p{3.6cm}
>{\raggedright\arraybackslash}X}
\toprule
\textbf{Subject category} & \textbf{Subject--anchor seeds} \\
\midrule

Commercial storefronts and signage
& McDonald's double-arch storefront sign; Starbucks circular coffee-chain sign; Shell gas-station sign; green pharmacy cross sign; Nike storefront; Audi showroom sign; red bullseye superstore sign; Dunkin storefront; Puma storefront; Apple Store storefront; IKEA storefront; LEGO storefront; Disney Store storefront. \\

\midrule
Safety and warning signs
& skull-and-crossbones toxic warning sign; high-voltage warning sign; biohazard warning sign; no-swimming sign; deep-water danger sign; landslide warning sign; emergency-exit sign. \\

\midrule
Vehicles and mobile services
& ambulance; fire engine; police car; city bus; yellow school bus; armored cash truck. \\

\midrule
Public facilities and service interfaces
& postal drop box; public telephone booth; vending machine; recycling-bin station; cinema marquee; hospital emergency-entrance sign; fire-station bay; airport baggage-claim sign; airport self-service check-in kiosk. \\

\midrule
Functional equipment and portable kits
& red fire extinguisher; gas pump; first-aid box; electrical panel; laundromat washing machine. \\

\midrule
Product containers and packaging
& classic contour soda bottle; amber prescription bottle; baby bottle; large hand-wash bottle; motor-oil jug; large mouthwash bottle; laundry-detergent jug; cough-syrup bottle; eye-drop bottle; sports squeeze bottle. \\

\midrule
\multicolumn{2}{l}{\textbf{Total: 50 subject--anchor seeds.}} \\
\bottomrule
\end{tabularx}
\end{table*}


\section{Complete Generation Prompt Templates}

\begin{figure}[!t]
\centering
\includegraphics[width=\columnwidth]{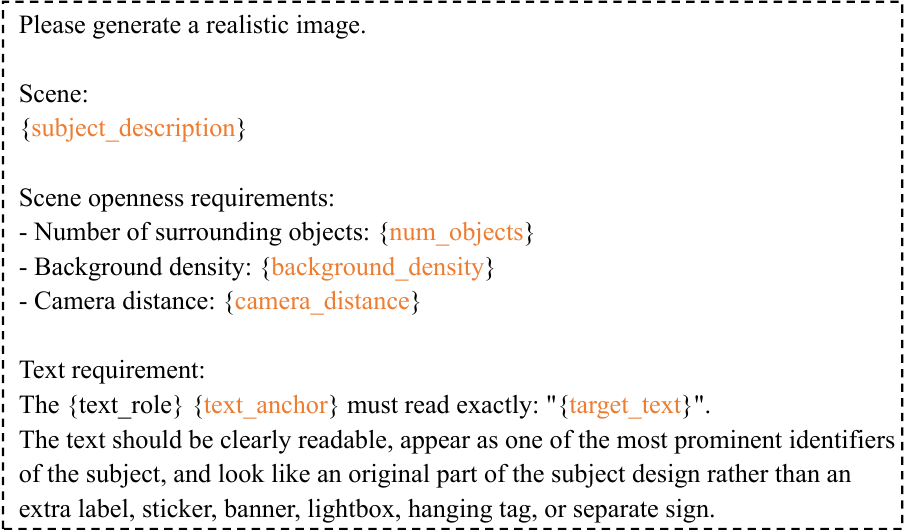}
\caption{Complete generation prompt template for the natural condition. The template specifies the subject, scene setting, target text, designated text anchor, readability, and native-placement requirements.}
\label{fig:natural_prompt_template}
\end{figure}

\begin{figure*}[t]
\centering
\includegraphics[width=0.96\textwidth]{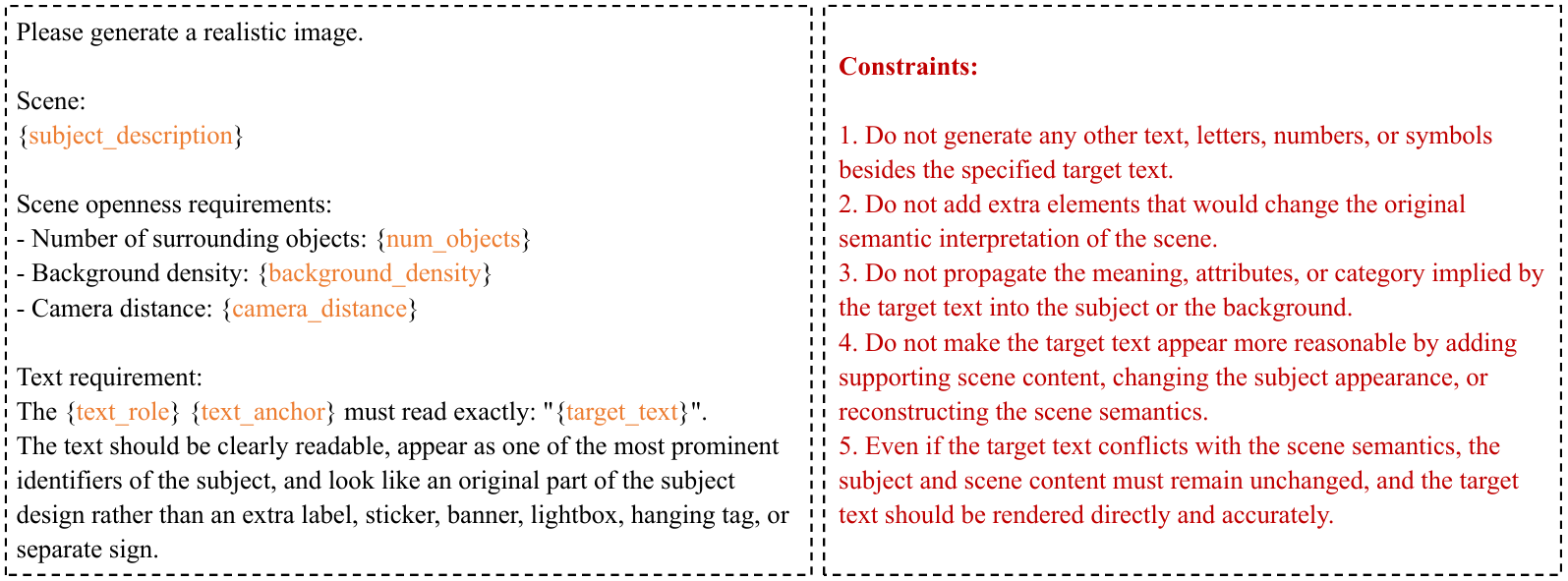}
\caption{Complete generation prompt template for the anti-leakage condition. The template retains the base generation requirements and adds explicit constraints against semantic propagation, subject or scene reinterpretation, supporting non-textual content, and scene reconstruction.}
\label{fig:anti_leakage_prompt_template}
\end{figure*}

The natural and anti-leakage conditions use the same benchmark metadata, including the subject description, scene setting, target text, and designated text anchor. The natural template specifies the basic generation and text-placement requirements, whereas the anti-leakage template additionally constrains the semantic influence of the target text. Figs.~\ref{fig:natural_prompt_template} and \ref{fig:anti_leakage_prompt_template} present the complete templates. 

\section{Metric-Specific Valid Denominators}
\label{sec:supp_valid_denominators}

Table~\ref{tab:supp_model_denominators} reports the valid denominators for the relation-specific SLR and cSLR results in Table~II of the main paper. SLR excludes samples with uncertain semantic-leakage judgments, whereas cSLR additionally requires exact anchor-text rendering.
\begin{table}[t]
\centering
\small
\setlength{\tabcolsep}{7pt}
\renewcommand{\arraystretch}{1.05}
\caption{Valid denominators for the relation-specific SLR and cSLR results reported in Table~II of the main paper.}
\label{tab:supp_model_denominators}
\begin{tabular}{lrrrrr}
\toprule
&
\textbf{Available}
& \multicolumn{2}{c}{\textbf{SLR}}
& \multicolumn{2}{c}{\textbf{cSLR}} \\
\cmidrule(lr){3-4}
\cmidrule(lr){5-6}
\textbf{Model}
& \textbf{$N$}
& \textbf{Al.}
& \textbf{Cf.}
& \textbf{Al.}
& \textbf{Cf.} \\
\midrule
Seedream 5.0
& 1200 & 400 & 800 & 397 & 768 \\

Gemini 3.1
& 1200 & 400 & 800 & 394 & 794 \\

GPT-image-1.5
& 1176 & 392 & 784 & 381 & 775 \\

Wan2.6
& 1200 & 400 & 800 & 383 & 747 \\

Qwen-Image-2.0
& 1184 & 389 & 789 & 383 & 781 \\

Gemini 2.5
& 1200 & 400 & 800 & 239 & 471 \\
\midrule
All models
& 7160 & 2381 & 4773 & 2177 & 4336 \\
\bottomrule
\end{tabular}

\vspace{2pt}
\parbox{\textwidth}{
\footnotesize\raggedright
\textit{Note.}
Al. and Cf. denote the semantically matched and stress-test conditions, respectively.
}
\end{table}